\documentclass{elsarticle}
\usepackage{fullpage}
\makeatletter
\def\ps@pprintTitle{%
\let\@oddhead\@empty
\let\@evenhead\@empty
\def\@oddfoot{\reset@font\hfil\thepage\hfil}
\let\@evenfoot\@oddfoot
}
\makeatother

\pdfoutput=1
\usepackage[utf8]{inputenc}
\usepackage{amsmath}
\usepackage{amssymb}
\usepackage{amsfonts}
\usepackage{amssymb}
\usepackage[pdftex]{hyperref}
\usepackage{graphicx}
\usepackage{breakcites}
\usepackage{color}
\usepackage{bm}
\usepackage{multirow}
\usepackage{booktabs}
\usepackage{soul}
\usepackage[raggedright]{sidecap}
\sidecaptionvpos{figure}{c} 
\usepackage{multirow}
\usepackage{floatrow}
\newfloatcommand{capbtabbox}{table}[][\FBwidth]

\biboptions{sort&compress}

\definecolor{darkgreen}{rgb}{0,0.6,0.2}
\definecolor{orange}{rgb}{1.0,0.5,0.0}

\usepackage{url}

\usepackage{subfigure}

\begin{document}

\title{The Deep Kernelized Autoencoder} 

\author[1]{Michael Kampffmeyer}
\author[1]{Sigurd L{\o}kse}
\author[1]{Filippo M. Bianchi}
\author[1,2]{Robert~Jenssen}
\author[3]{Lorenzo Livi}
\address[1]{Machine Learning Group, UiT--The Arctic University of Norway, http://site.uit.no/ml/}
\address[2]{Norwegian Computing Center, Oslo, Norway}
\address[3]{Department of Computer Science, University of Exeter, UK}

\begin{abstract}
Autoencoders learn data representations (codes) in such a way that the input is reproduced at the output of the network.
However, it is not always clear what kind of properties of the input data need to be captured by the codes.
Kernel machines have experienced great success by operating via inner-products in a theoretically well-defined reproducing kernel Hilbert space, hence capturing topological properties of input data.
In this paper, we enhance the autoencoder's ability to learn effective data representations by aligning inner products between codes with respect to a kernel matrix.
By doing so, the proposed \emph{kernelized} autoencoder allows learning similarity-preserving embeddings of input data, where the notion of similarity is explicitly controlled by the user and encoded in a positive semi-definite kernel matrix.
Experiments are performed for evaluating both reconstruction and kernel alignment performance in classification tasks and visualization of high-dimensional data. Additionally, we show that our method is capable to emulate kernel principal component analysis on a denoising task, obtaining competitive results at a much lower computational cost.
\end{abstract}

\maketitle

\section{Introduction}
\label{sec:intro}

Autoencoders (AEs) are a class of neural networks that gained increasing interest in recent years~\cite{vincent2010stacked,kingma2013auto, santana2016information,tolstikhin2017wasserstein,2018adversarially}.
AEs are used for unsupervised learning of \emph{effective} latent representations of data ~\cite{Hinton504,bengio2009learning}.
However, what an \emph{effective} representation consists of is highly dependent on the target task, such as clustering and classification \cite{6472238}.
In standard AEs, representations are derived by training the network to reconstruct inputs through either a bottleneck layer, thereby forcing the network to learn how to compress inputs, or through an over-complete representation. It can be shown that training autoencoders using a reconstruction error corresponds to maximizing the lower bound of the mutual information between input and the learned representation~\cite{vincent2010stacked}.
Regularization methods are commonly employed for enforcing sparseness, improving robustness to noise, avoiding trivial identity mappings, or penalizing sensitivity of the representation to small changes in inputs~\cite{6472238}.
Nonetheless, regularization alone provides limited control over the nature of the hidden representation.

In this paper, we propose a method to learn representations that preserve desired similarities in input space with an AE.
In our approach, similarities are encoded in form of a kernel matrix, which is used as a prior to be reproduced by inner products of the hidden representations learned by the AE. 
This allows us to learn data representations with specified pairwise relationships.
The training loss minimizes a combination of reconstruction error and a term quantifying the misalignment of the prior and the inner products of the hidden representations; the misalignment is computed by means of the normalized Frobenius norm.
We note that this process acts as a regularization for the hidden representations and resembles the well-known kernel alignment procedure \cite{wang2015overview}.
Our contribution is in principle related to other well-established methods like those from the family of multidimensional scaling~\cite{bronstein2006generalized}, where an explicit embedding of the data is computed by minimizing a measure of distortion based on inner products.
Further, we will experimentally show that the proposed regularization method allows mitigating a problem often observed in non-regularized AEs, where codes for similar images are not similar themselves and the underlying manifold is disconnected~\cite{makhzani2015adversarial}.

\subsection{Related Works}
The proposed model, called \textit{deep kernelized autoencoder}, is related to recent attempts to 
incorporate kernel and information theoretic learning methods within neural network architectures~\cite{wilson2016deep, NIPS2009_3628}. 
Specifically, it is connected to works on interpreting neural networks from a kernel perspective~\cite{Montavon2011} and the Information Theoretic-Learning Auto-Encoder~\cite{santana2016information}, which imposes a prior distribution over the hidden representation in a variational autoencoder~\cite{kingma2013auto}. 
Achille and Soatto~\cite{achille2016information} proposed a regularization method exploiting information dropout, an information-theoretic generalization of dropout~\cite{srivastava2014dropout} for neural networks and show that an AE trained with such a regularization for a specific parameter setting simplifies to the variational autoencoder objective. 
Other information-theoretic learning concepts, such as the information bottleneck~\cite{still2014information}, have also recently emerged in the deep learning literature \cite{shwartz2017opening}.
In~\cite{alemi2016deep} variational inference is used to optimize the lower bound on the information bottleneck to learn representations that maximize the mutual information between learned representation and output while minimizing the mutual information between input and hidden representation. 
Computing the information bottleneck is difficult, especially with high-dimensional data. \citet{NIPS2016_6101} proposed an efficient variational scheme for maximizing a lower bound of the original information bottleneck formulation, which also allows for non-linear mappings between input and compressed representation via kernel functions.
Beside dimensionality reduction, neural networks utilizing kernel and information theoretic concepts have also been used to perform clustering~\cite{kampffmeyer2017clustering}.

In our work, we exploit kernel alignment to match the inner products of the learned representations with a similarity measure in the input space encoded as a kernel matrix.
A recent related work in this direction by~\citet{horn2017learning} attempts to learn representations that preserve pairwise similarity by means of AEs.
The authors specifically focus on dimensionality reduction, showing the possibility to approximate the pairwise data similarity in input space in linear fashion from the learned low-dimensional representation. In practice, given an input data point, the network is trained to recreate the related row of the similarity matrix. Recently, \citet{chu2017stacked} propose a similarity-preserving AE based on clustering data in input space. Hidden representations are learned in such a way that data points belonging to the same cluster are similar also in the hidden representation.

Another recent approach consists in integrating Wasserstein Generative Adversarial Neural Networks into the AE framework~\cite{2018adversarially}. Similarly,~\citet{tolstikhin2017wasserstein} proposes the Wasserstein Autoencoder, which is based on a novel regularization technique minimizing the Wasserstein distance between the model distribution and a target distribution.


\subsection{Contribution and paper organization}
In addition to providing more control over hidden representations, our method also has several benefits that compensate for important drawbacks of traditional kernel methods.
By means of an end-to-end training procedure, we learn an explicit approximate mapping function from the input to a kernel space, as well as the associated back-mapping to the input space.
Once the mapping is learned, it can be applied to inputs and operations performed in kernel space can then be explicitly simulated by means of linear operations in code space, thus in practice allowing to perform non-linear operations in input space.
Mini-batch training is used in the proposed method in order to lower the computational complexity inherent to traditional kernel methods and, especially, spectral methods~\cite{scholkopf1998nonlinear,boser1992training,jenssen2010kernel}.
Furthermore, our method can be used with arbitrary kernel functions, even those computed with an algorithmic procedure, i.e., where inner products in kernel space are not expressed by an analytic function.
To stress this fact, in our experiments we consider the probabilistic cluster kernel (PCK), a kernel function that is the result of a feature generation procedure. PCK is robust with respect to hyperparameter choices and has been shown to often outperform counterparts such as the radial basis function (RBF) kernel~\cite{izquierdo2015spectral}.

A preliminary version of this method appeared in~\cite{kernelAE}. Here we extend our work by:
\begin{itemize}
    \item providing a thorough literature background discussion, placing our work into a broader context;
    \item extending the experimental evaluation to additional datasets, namely (i) the image dataset CIFAR-10, (ii) the text dataset Reuters, and (iii) the remote sensing dataset Cloud;
    \item experimentally analyzing the effectiveness of the learned representations for classification tasks and visualizing high-dimensional data, and for generating new data samples beyond those seen during training.
\end{itemize}

The paper is structured as follows. Section~\ref{sec:background} provides the reader with a discussion of the relevant background, such as AEs and kernel methods; notably in Section \ref{sec:pck} we introduce PCK, adopted here for obtaining kernel matrices to be used in our method. Section~\ref{sec:kernelizedae} describes the proposed methodology. Experimental results are discussed in Section~\ref{sec:analysis} and Section~\ref{sec:experiments}. Finally, Section~\ref{sec:conclusions} draws conclusions and points to future research directions.

\section{Background}
\label{sec:background}

\subsection{Autoencoders and stacked autoencoders}
\label{sec:autoencoders}

AEs simultaneously learn two functions.
The first one, the \textit{encoder}, provides a mapping from an input domain, $\mathcal{X}$, to a code domain, $\mathcal{C}$, i.e., the hidden representation.
The second function, the \textit{decoder}, maps from $\mathcal{C}$ back to $\mathcal{X}$.
For a single hidden layer AE, the encoding function $E(\cdot)$ and the decoding function $D(\cdot)$ are defined as
\begin{align}
    \begin{split}
    \label{eq:encoding_decoding}
    \mathbf{h} &= E(\mathbf{x}) = \sigma(\mathbf{W}_E\mathbf{x} + \mathbf{b}_E) \\
    \mathbf{\tilde{x}} &= D(\mathbf{h}) = \sigma(\mathbf{W}_D\mathbf{h} + \mathbf{b}_D),
    \end{split}
\end{align}
where $\sigma(\cdot)$ denotes a suitable transfer function (e.g., a sigmoid applied component-wise), $\mathbf{x}$, $\mathbf{h}$, and $\mathbf{\tilde{x}}$ denote, respectively, a sample from the input space, its hidden representation also called \emph{code}, and its reconstruction; finally, $\mathbf{W}_{E}$ and $\mathbf{W}_{D}$ are the weights, and $\mathbf{b}_{E}$ and $\mathbf{b}_{D}$ the bias of encoder and decoder, respectively.

In order to minimize the discrepancy between the original data and its reconstruction, model parameters in Equation \ref{eq:encoding_decoding} are learned by minimizing, usually through stochastic gradient descent (SGD), a reconstruction loss of the form
\begin{equation}
    \label{eq:distortion}
    L_r(\mathbf{x}, \mathbf{\tilde{x}}) = \lVert \mathbf{x} - \mathbf{\tilde{x}} \rVert_{2}^{2} \; .
\end{equation}
%

Differently from Equation~\ref{eq:encoding_decoding}, a stacked autoencoder (sAE) consists of several hidden layers \cite{Hinton504}. 
Deep architectures are capable of learning complex representations by transforming input data through multiple layers of nonlinear processing \cite{6472238}.
The optimization of the weights is harder in this case and pretraining is beneficial, as it is often easier to learn intermediate representations, instead of training the whole architecture end-to-end~\cite{bengio2009learning}. A 
common application of pre-trained sAE is the initialization of layers in deep neural networks~\cite{vincent2010stacked}.
Pretraining is performed in different phases, each of which consists of training a single AE layer. After the first AE has been trained, its encoding function $E(\cdot)$ is kept fixed and is applied to the input and the resulting representation is used to train the next AE in the stacked architecture. Each layer, being trained independently, aims at capturing more abstract features by trying to reconstruct the representation in the previous layer. Once all individual AEs are trained, their hidden layers (encoding and decoding functions) are extracted and stacked on each other, yielding a pre-trained sAE.

\subsection{A brief introduction to kernel methods}
\label{sec:kernel_methods}
Kernel methods process data in a reproducing kernel Hilbert space (RKHS) $\mathcal{K}$ associated with an input space $\mathcal{X}$ through an implicit (non-linear) mapping $\phi: \mathcal{X} \rightarrow \mathcal{K}$.
There, data are more likely to become separable by linear methods~\cite{cover1991elements}, which produces results that are otherwise only obtainable by nonlinear operations in the input space.
Explicit computation of the mapping $\phi(\cdot)$ and its inverse $\phi^{-1}(\cdot)$ is, in practice, not required. In fact, operations in the kernel space are expressed through inner products (kernel trick), which are computed as Mercer kernel functions in input space: $\kappa(\mathbf x_i, \mathbf x_j) = \langle \phi(\mathbf x_i), \phi(\mathbf x_j) \rangle$.

As a major drawback, kernel methods scale poorly with the number of samples $n$: traditionally, memory requirements of these methods scale with $\mathcal{O}(n^2)$ and computation with $\mathcal{O}(n^2\times d)$, where $d$ is the input dimension~\cite{dai2014scalable}.
For example, kernel principal component analysis (kPCA)~\cite{scholkopf1998nonlinear}, a common dimensionality reduction technique that projects data into the subspace that preserves the maximal amount of variance in kernel space, requires to compute the eigendecomposition of a kernel matrix $\mathbf{K} \in \mathbb{R}^{n\times n}$, with $K_{ij}=\kappa(x_i, x_j), x_i, x_j\in\mathcal{X}$, with computational and memory costs scaling as $\mathcal{O}(n^3)$ and $\mathcal{O}(n^2)$, respectively. For this reason, kPCA is not applicable to large-scale problems.
The availability of efficient (approximate) mapping functions, however, would reduce the complexity, thereby enabling these methods to be applicable to larger datasets~\cite{NIPS2009_3628}. In this direction, \citet{NIPS2007_3182} and \citet{6136519} proposed approximate mappings preserving the dot product structure by using low-dimensional randomized features, hence allowing the use of fast linear methods in an explicit way.
Furthermore, finding an explicit inverse mapping from $\mathcal{K}$ to the input domain is a central problem in several applications, such as image denoising performed with kPCA, also known as the pre-image problem~\cite{bakir2004learning,honeine2011closed}.

Our proposed method instead, attempts to approximate the operations in the kernel space using an AE architecture that scales to large datasets, provides an implicit inverse mapping, and, once trained, can process new samples efficiently.

\subsection{Probabilistic cluster kernel}
\label{sec:pck}
The Probabilistic Cluster Kernel (PCK)~\cite{izquierdo2015spectral} is a robust
    kernel function, which automatically adapts to the inherent structures in the data.
Its robustness comes from the fact that it does not depend on any critical user--specified
    hyperparameters, like the width in Gaussian kernels.
The PCK is trained by fitting multiple Gaussian Mixture Models (GMMs) to
    the input data using the EM algorithm and combining these models to
    generate a single kernel.
In particular, GMMs are trained using different number of mixture components
    $g = 2, 3, \ldots, G$, each with different randomized initial
    conditions $q = 1, 2, \ldots, Q$.
Let $\boldsymbol \pi_i(q, g)$ denote the \textit{posterior distribution} for
    data point $\mathbf x_i$ under a GMM with $g$ mixture components and initial
    condition $q$.
The PCK is then defined as
\begin{equation}\label{eq:pck}
    \kappa_{\mathrm{PCK}}(\mathbf x_i, \mathbf x_j) 
    = \frac{1}{Z} \sum_{q = 1}^Q \sum_{g = 2}^G
    \boldsymbol \pi_i^T(q, g) \boldsymbol \pi_j (q, g),
\end{equation}
where $Z$ is a normalizing constant.

Intuitively, the posterior distribution under a mixture model contains probabilities
    that a given data point belongs to a certain mixture component in the model.
Thus, the inner products in Equation~\ref{eq:pck} are large if data pairs often belong
    to the same mixture component.
By averaging these inner products over a range of $g$ values, the kernel function has
    a large value if these data points are similar on both global scale
    (small $g$ $\rightarrow$ large mixture components) and local scale (large $g$ $\rightarrow$ small mixture components).
    
The PCK has previously been used for semi-supervised learning \cite{izquierdo2014semisupervised}
    and spectral clustering \cite{izquierdo2015spectral}.
Additionally, variations of the method for handling missing data have been proposed for both time series
    \cite{mikalsen2018time} and vectorial data \cite{lokse2017spectral}.

\section{Deep kernelized autoencoders}
\label{sec:kernelizedae}
\begin{figure}[t!]
  \centering
\includegraphics[width=0.6\textwidth, keepaspectratio]{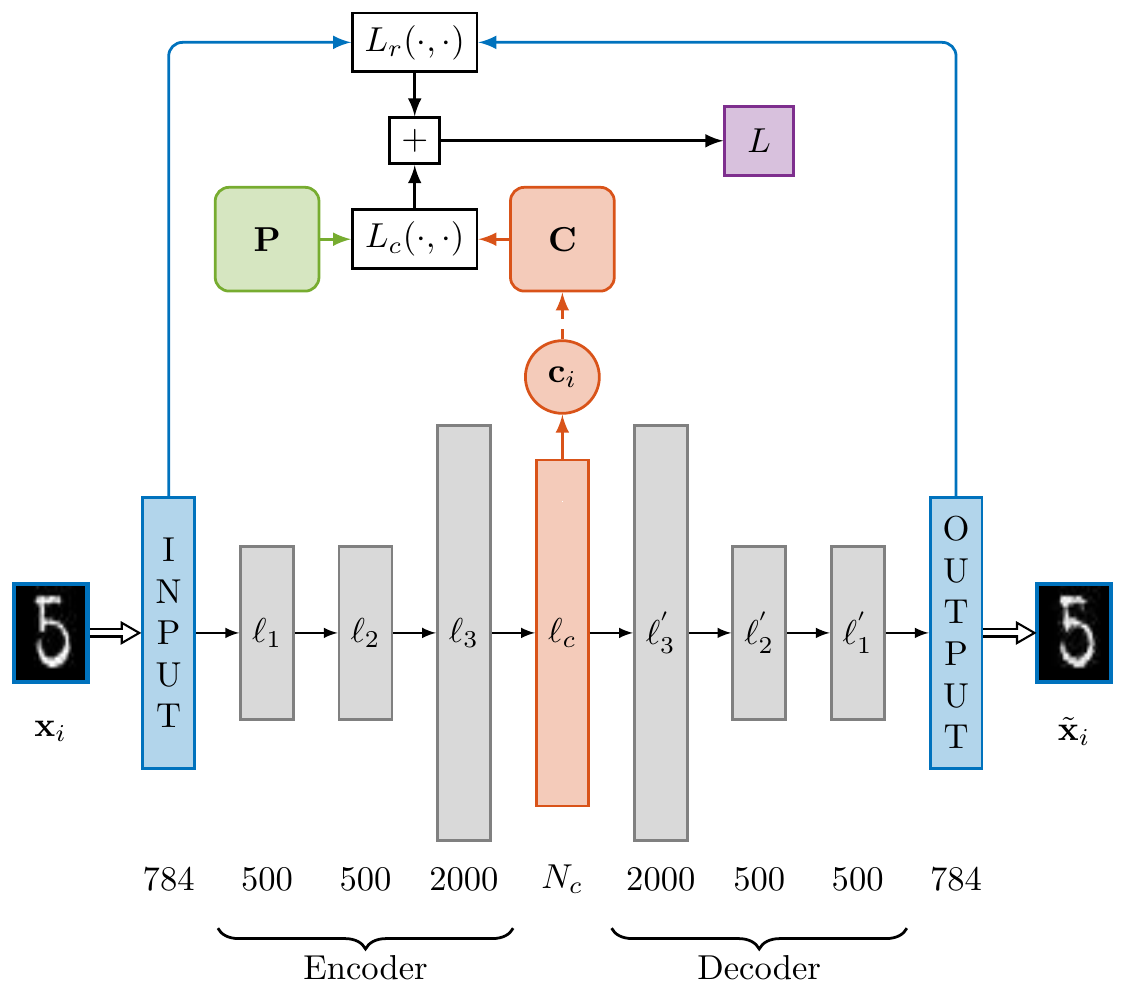}
  \caption{Schematic illustration of dkAE architecture. Loss function $L$ depends on two terms. First, $L_r(\cdot,\cdot)$, is the reconstruction error between the true input $\mathbf{x}_i$ and the output of the dkAE, $\tilde{\mathbf{x}}_i$. The second term, $L_c(\cdot, \cdot)$, is the distance measure between matrices $\mathbf{C}$ (computed as inner products of codes $\{ \mathbf{c}_i \}_{i=1}^{n}$) and the target prior kernel matrix $\mathbf{P}$. For mini-batch training the matrix $\mathbf{C}$ is computed over the codes of the data in the mini-batch and that distance is compared to the submatrix of $\mathbf{P}$ related to the current mini-batch.}
    \label{fig:kAE_arch}
\end{figure}

In this section, we describe our contribution, which is a method combining deep AEs with kernel methods: the deep kernelized AE (dkAE).
A dkAE is trained by minimizing the following loss function
\begin{equation}
    \label{eq:cost}
    L = (1-\lambda) L_r(\mathbf{x}, \mathbf{\tilde{x}}) + \lambda L_c(\mathbf{C}, \mathbf{P}),
\end{equation}
where $L_r(\cdot, \cdot)$ is the reconstruction loss in Equation~\ref{eq:distortion}.
$L_c(\cdot, \cdot)$ is the code loss, a distance measure between two matrices, $\mathbf{P} \in \mathbb{R}^{n \times n}$, the kernel matrix given as prior, and $\mathbf{C} \in \mathbb{R}^{n \times n}$, the inner product matrix of codes associated to the input data.
The objective of $L_c(\cdot,\cdot)$ is to enforce the similarity between $\mathbf{C}$ and the kernel matrix $\mathbf{P}$.
$\lambda$ is a hyperparameter ranging in $[0, 1]$, which weights the importance of the two objectives in Equation~\ref{eq:cost}; for $\lambda=0$, the loss function simplifies to the traditional AE loss in Equation~\ref{eq:distortion}.
A depiction of the training procedure is reported in Figure~\ref{fig:kAE_arch}.

We implement $L_c(\cdot, \cdot)$ as the normalized Frobenius distance between $\mathbf{C}$ and $\mathbf{P}$. Each matrix element $C_{ij}$ in $\mathbf{C}$ is given by 
$C_{ij}=E(\mathbf{x}_i) \cdot E(\mathbf{x}_j)$ and the code loss is computed as
\begin{equation}
\label{eq:regularization}
    L_c(\mathbf{C}, \mathbf{P}) = \Bigg{\lVert} \frac{\mathbf{C}}{\|\mathbf{C}\|_F} - \frac{\mathbf{P}}{\|\mathbf{P}\|_F} \Bigg{\rVert}_{F}.
\end{equation}
    
It is worth noting that minimizing the normalized Frobenius distance between the kernel matrices is equivalent to maximizing the traditional kernel alignment cost, since
\begin{equation} \label{eq:normdistance}
   \Bigg{\lVert} \frac{\mathbf{C}}{\|\mathbf{C}\|_F} - \frac{\mathbf{P}}{\|\mathbf{P}\|_F} \Bigg{\rVert}_{F} =
   \sqrt{2 - 2A(\mathbf C, \mathbf P)},
\end{equation} 
where $A(\mathbf C, \mathbf P) = \frac{\langle \mathbf C, \mathbf P \rangle_F}
{\|\mathbf C\|_F \|\mathbf P\|_F}$ is exactly the kernel alignment cost function \cite{christianini2001kernel,wang2015overview}.
Note that the distance in Equation \ref{eq:normdistance} can be implemented also with more advanced differentiable measures of (dis)similarity between positive-definite matrices, such as divergence and mutual information \cite{kulis2009low,giraldo2015measures}. However, these options are not explored in this paper and are left for future research.

In this paper, the prior kernel matrix $\mathbf{P}$ is computed by means of the PCK algorithm introduced in Section \ref{sec:pck}, such that $\mathbf{P}=\mathbf{K}_{\text{PCK}}$. 
However, our approach is general and \textit{any} kernel matrix can be used as prior in Equation~\ref{eq:regularization}.

Note, that the kernel alignment also acts as a regularization, discouraging the learning of trivial mappings. Furthermore, we also employ tied weights in the encoder and decoder as additional regularization following~\cite{kamyshanska2015potential}.

\subsection{Mini-batch training}
We use mini batches of $k$ samples to train the dkAE, thereby avoiding the computational restrictions of kernel and especially spectral methods outlined in Section~\ref{sec:kernel_methods}.
In particular, the memory complexity of the algorithm can be reduced to $\mathcal{O}(k^2)$, where $k \ll n$. Finally, we note that the computational complexity scales linearly with regards to the network parameters.
Given a mini batch of $k$ samples, the dkAE loss function is defined by taking the average of the per-sample reconstruction cost
\begin{equation}
\label{eq:cost_minibatch}
    L_{\mathrm{batch}}=\frac{1-\lambda}{kd} \sum_{i=1}^{k} L_r(\mathbf{x}_i, \mathbf{\tilde{x}}_i)  + \lambda
   \Bigg{\lVert} \frac{\mathbf{C}_k}{\|\mathbf{C}_k\|_F} - \frac{\mathbf{P}_k}{\|\mathbf{P}_k\|_F} \Bigg{\rVert}_{F},
\end{equation}
where $d$ is the dimensionality of the input space, $\mathbf{P}_k$ is a subset of $\mathbf{P}$ that contains only the $k$ rows and columns related to the current mini-batch, and $\mathbf{C}_k$ contains the inner products of the codes related to the mini-batch. Note that $\mathbf{C}_k$ is re-computed for each mini batch ($\mathcal{O}(k^2)$), while $\mathbf{P}_k$ is obtained by means of indexing operations with cost $\mathcal{O}(k)$.

\subsection{Operations in code space}
\label{sec:lin_op}
\begin{figure}[tbp]
  \centering
  \includegraphics[width=0.5\textwidth, keepaspectratio]{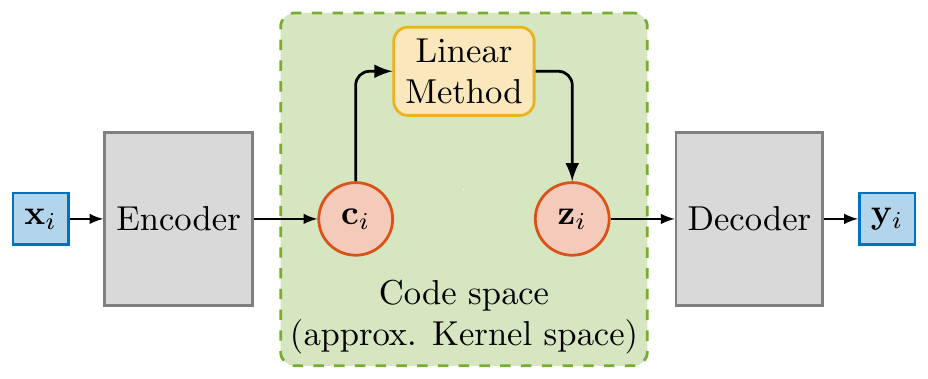}
  \caption{The encoder maps input $\mathbf{x}_i$ to $\mathbf{c}_i$, which lies in code space. In dkAEs, the code domain approximates the space associated to the prior kernel $\mathbf{P}$. A linear method receives input $\mathbf{c}_i$ and produces output $\mathbf{z}_i$.
  The decoder maps $\mathbf{z}_i$ back to input space. The result $\mathbf{y}_i$ can be seen as the output of a non-linear operation on $\mathbf{x}_i$ in input space.}
    \label{fig:kAE_method}
\end{figure}

Linear operations in code space can be performed as shown in Figure \ref{fig:kAE_method}.
The encoding scheme of the proposed dkAE implicitly approximates $\phi(\cdot)$, mapping an input $\mathbf{x}_i$ onto the kernel space.
In particular, in dkAEs, the feature vector $\phi(\mathbf{x}_i)$ is approximated by the code $\mathbf{c}_i$. Our non-linear encoder maps the inputs into a space where they are more likely to be linearly separable, as there the code vectors preserve a non-linear similarity computed in the input space.
A linear operation on $\mathbf{c}_i$ produces a result in the code space, $\mathbf{z}_i$, relative to the input $\mathbf{x}_i$.
Unlike other kernel methods where the explicit mapping back to the input space is not defined, we can map codes back by means of a decoder, which in our case approximates the inverse mapping $\phi(\cdot)^{-1}$ from the kernel space back to the input domain. 
This enables dkAEs to provide visualization and interpretation of the results in the original space; we further explore these perspectives in the experiments.

\section{Analysis of dkAE}
\label{sec:analysis}
In this section, we perform an analysis of the proposed method by considering three experiments.
Section \ref{sec:exp-setting} delineates the experimental setting.
In Section \ref{sec:exp1}, we evaluate the sensitivity of the two terms in the objective function (Equation \ref{eq:cost_minibatch}) when varying the $\lambda$ hyperparameter (in Equation \ref{eq:cost}) and the size of the code layer (i.e., number of neurons in the innermost hidden layer).
Successively, in Section \ref{sec:exp1b} we evaluate the reconstruction accuracy and kernel alignment performance implemented by dkAEs.
Further, in Section \ref{sec:exp2} we compare dkAEs approximation accuracy of the prior kernel matrix with kPCA as the number of retained principal components increases.

\subsection{Experimental setting}
\label{sec:exp-setting}
The analysis is performed on the MNIST dataset, which consists of $60000$ handwritten digit images~\cite{lecun1998gradient}.
We use a subset of $20000$ samples due to the computational restrictions imposed by the PCK, which we use to illustrate dkAEs ability to learn arbitrary kernels, even if they originate from an ensemble procedure.

We train PCK by fitting GMMs on a subset of 200 training samples using parameters $Q=G=30$. These parameters are sufficiently large to ensure robust results~\cite{lokse2017spectral}.
Once trained, the GMM models are applied to the remaining data to calculate the whole kernel matrix.
We use 70\%, 15\% and 15\% of the data for training, validation, and testing, respectively.

The network architecture used in the experiments is $d-500-500-2000-N_c$ (see Figure \ref{fig:kAE_arch}), which has been demonstrated to perform well on several datasets, including MNIST, for both supervised and unsupervised tasks~\cite{maaten2009learning,hinton2006fast}.
Here, $N_c$ refers to the dimensionality of the code layer. Training was performed using the sAE pretraining approach outlined in Section~\ref{sec:autoencoders}.
To avoid learning the identity mapping on each individual layer, we applied a common \cite{kamyshanska2015potential} regularization technique where the encoder and decoder weights are tied, i.e., $W_{E} = W_{D}^T$. This is done during pretraining and fine-tuning. 
Unlike in traditional sAEs, to account for the kernel alignment objective, the code layer is optimized according to Equation~\ref{eq:cost} \textit{also} during pretraining.

Size of mini-batches for training was chosen to be $k=200$ randomly, independently sampled data points; in our experiments, an epoch consists of processing $(n/k)^2$ batches.
Pretraining is performed for $30$ epochs per layer and the final architecture is fine-tuned for $100$ epochs using gradient descent based on Adam~\cite{kingma2014adam}. 
The dkAE weights are randomly initialized according to Glorot et al.~\cite{Glorot10understandingthe}.

\subsection{Sensitivity analysis of hyperparameter $\lambda$ and size $N_c$ of code layer}
\label{sec:exp1}
\begin{figure}[tbp]
\centering
\subfigure[]{
\includegraphics[width=0.45\textwidth]{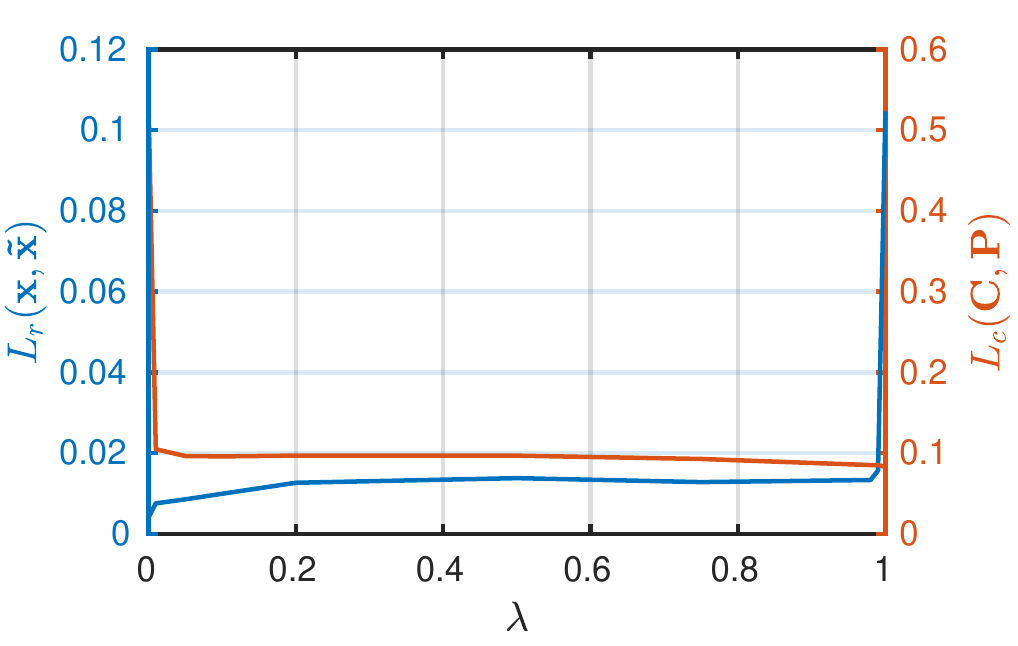}
\label{fig:lambda_experiment}}\hspace{-1em}
\subfigure[]{
\includegraphics[width=0.45\textwidth]{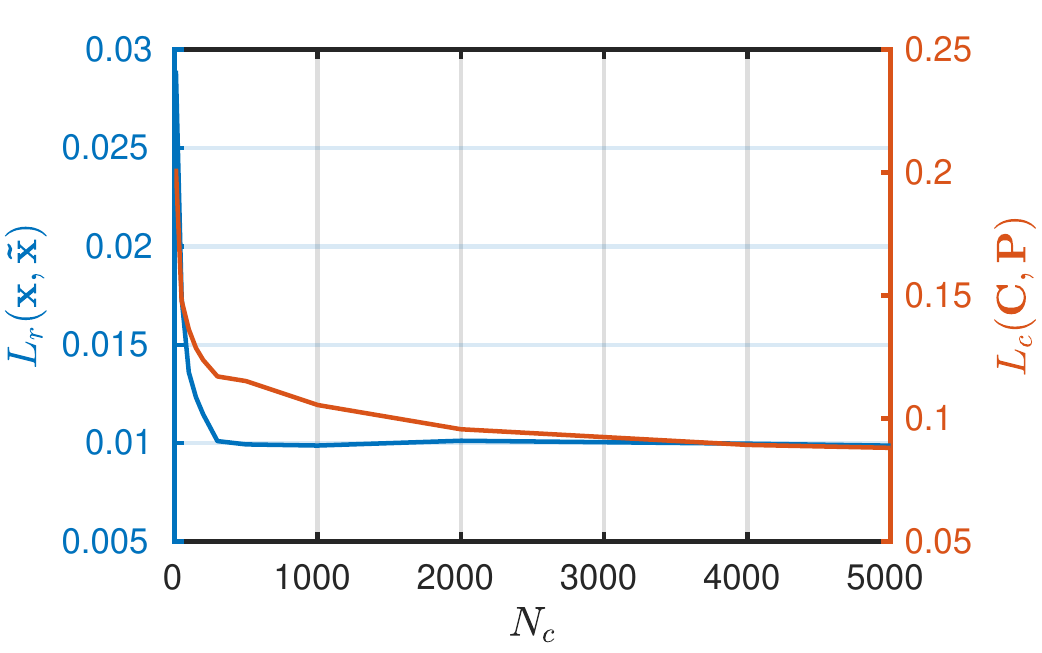}
\label{fig:nc_experiment}}\hspace{-1em}
\caption{(a): Tradeoff when choosing $\lambda$. High $\lambda$ values result in low $L_c$, but high reconstruction cost, and vice-versa. (b): Both $L_c$ and reconstruction costs decrease when code dimensionality $N_c$ increases.}
\label{fig:mnist_map}
\end{figure}

Here, we evaluate the influence of the two main hyperparameters influencing the resulting model.
Note that the experiments shown in this section are performed by training the dkAE on the training set and evaluating the performance on the validation set. We evaluate both the out-of-sample reconstruction $L_r$ and $L_c$. This is done in order to select the optimal parameters for evaluating the test set in the successive experiments.
Figure~\ref{fig:lambda_experiment} illustrates the effect of $\lambda$ for a fixed value $N_c=2000$ of neurons in the code layer. It can be observed that the reconstruction loss $L_r$ increases as more and more focus is put on minimizing $L_c$ (obtained by increasing $\lambda$). This quantifies empirically the trade-off in optimizing the reconstruction performance and the kernel alignment at the same time.
By inspecting the results, specifically the near constant losses for $\lambda$ in range $[0.1,0.9]$ the method appears robust to changes in hyperparameter $\lambda$.

Analyzing the effect of varying $N_c$ given a fixed $\lambda=0.1$ (Figure~\ref{fig:nc_experiment}), we observe that both losses decrease as $N_c$ increases. 
This could suggest that an even larger architecture, characterized by more layers and more neurons w.r.t. the architecture adopted here might work well, as the dkAE does not seem to overfit; due also to the regularization effect provided by the kernel alignment.

\subsection{Reconstruction error and kernel alignment}
\label{sec:exp1b}
\begin{figure}[tbp]
\centering
\minipage[t]{0.22\textwidth}
  \includegraphics[width=\linewidth]{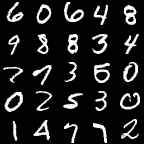}
  \centering{Original}
\endminipage\hspace{0.2cm}
\minipage[t]{0.22\textwidth}%
  \includegraphics[width=\linewidth]{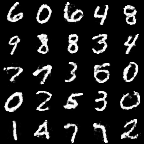}
  \centering{$\lambda=0.75$}
\endminipage\hspace{0.2cm}
\minipage[t]{0.22\textwidth}
  \includegraphics[width=\linewidth]{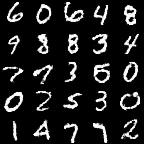}
  \centering{$\lambda=0.1$}
\endminipage\\
\minipage[t]{0.22\textwidth}
  \includegraphics[width=\linewidth]{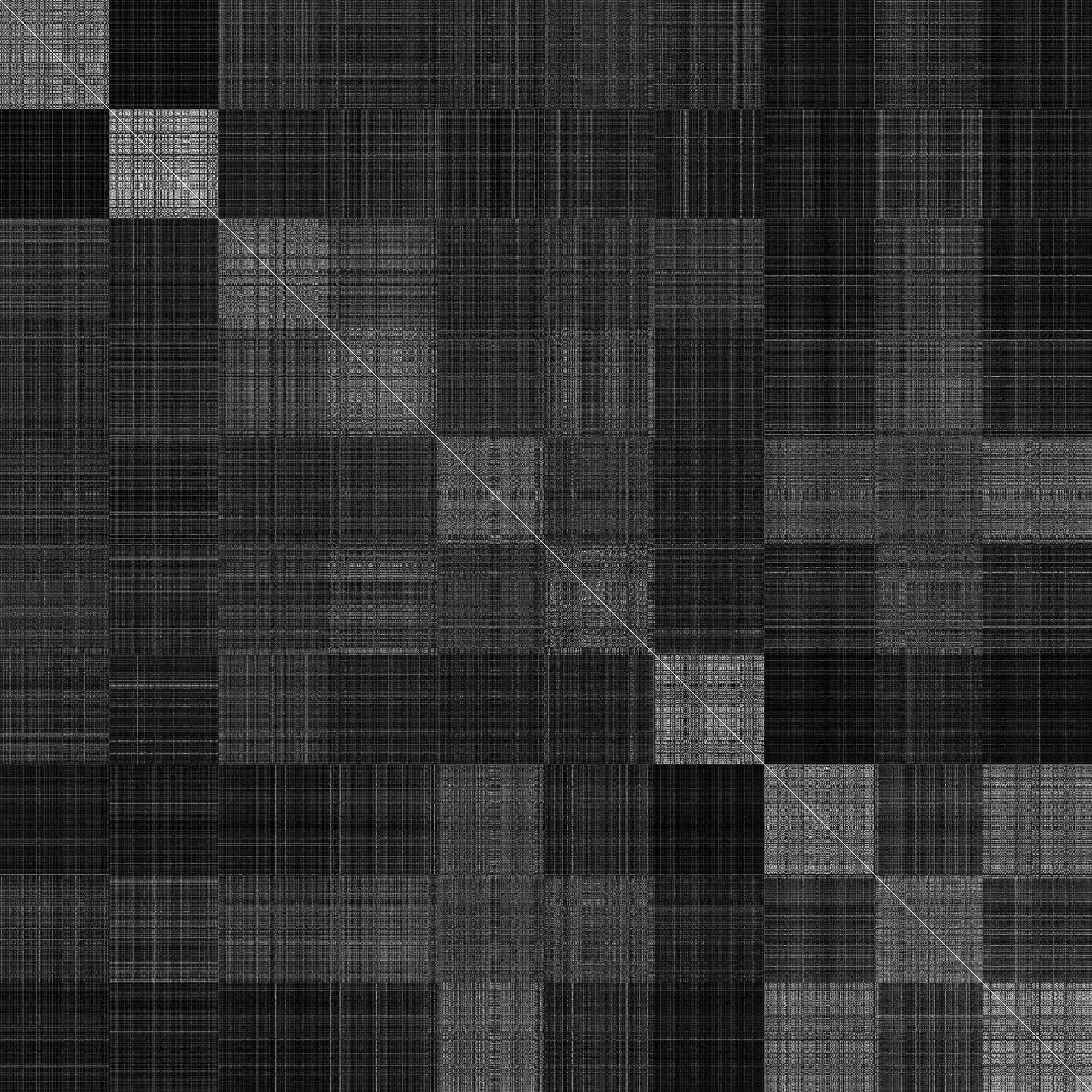}
  \centering{Prior}
\endminipage\hspace{0.2cm}
\minipage[t]{0.22\textwidth}
  \includegraphics[width=\linewidth]{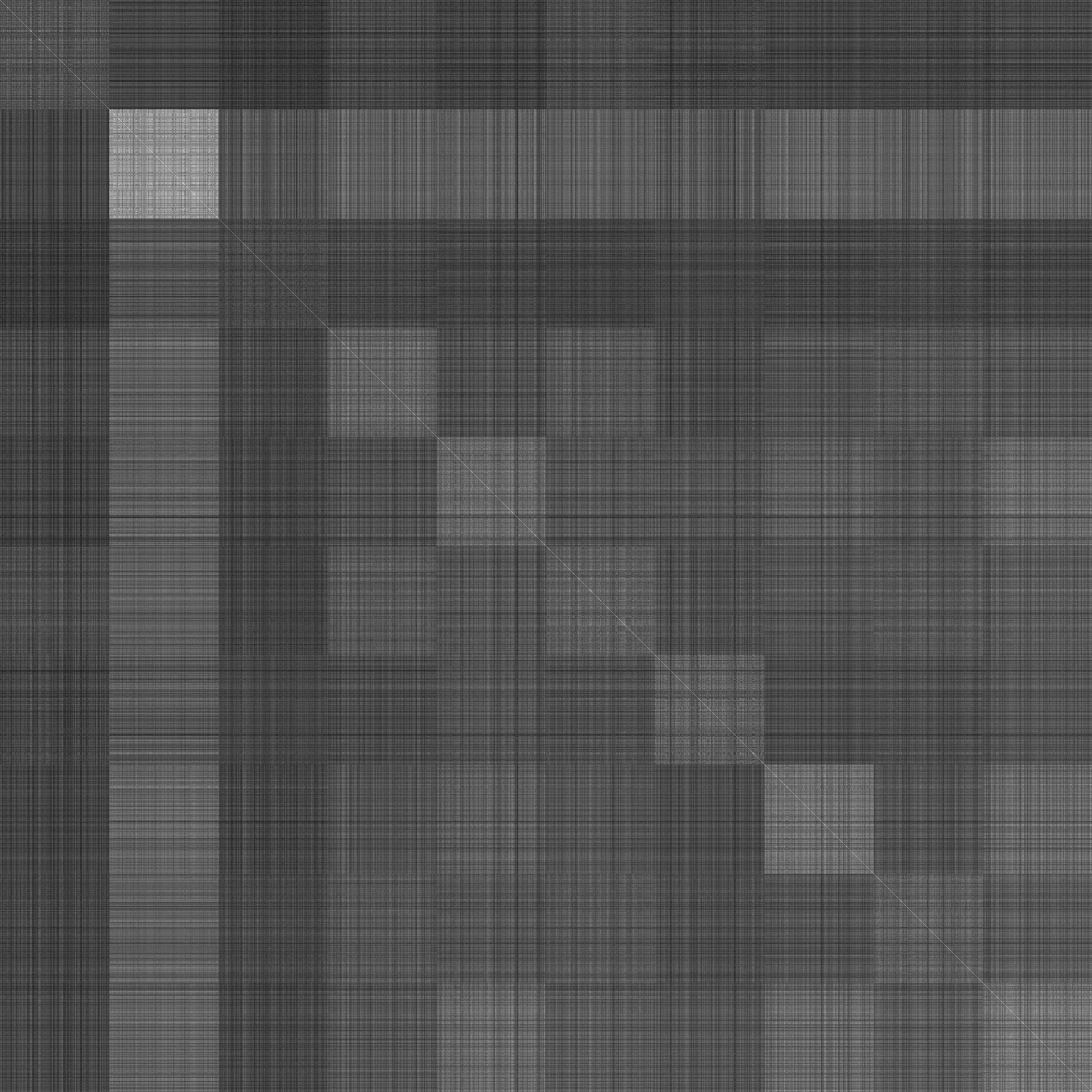}
  \centering{$\lambda=0.0$}
\endminipage\hspace{0.2cm}
\minipage[t]{0.22\textwidth}%
  \includegraphics[width=\linewidth]{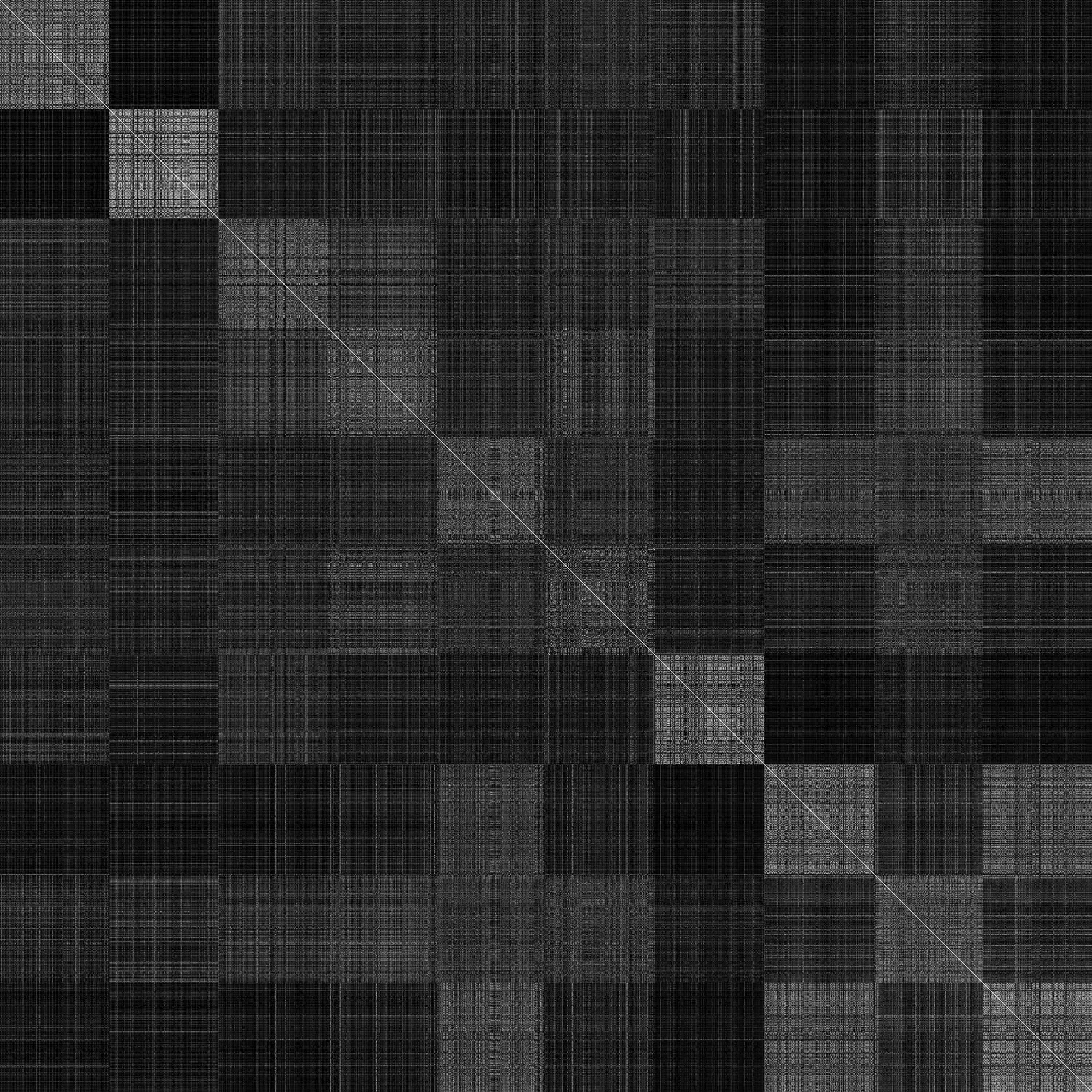}
  \centering{$\lambda=0.1$}
\endminipage
\caption{Illustrating the reconstruction error and kernel alignment trade-off in for different $\lambda$ values. We note that the reconstruction for a small $\lambda$ is generally better (see also Figure~\ref{fig:lambda_experiment}), but that small $\lambda$ yields high $L_c$.}\label{fig:lambda_experiment_reconstruction}
\end{figure}

By considering the previous results, in the following experiments we set $\lambda=0.1$ and $N_c=2000$.
Figure~\ref{fig:lambda_experiment_reconstruction} illustrates the results in Section~\ref{sec:exp1} qualitatively by displaying a set of original images from our test set and their reconstruction error for the chosen $\lambda$ value and a non-optimal one. 
Similarly, the prior kernel (rows/columns sorted by class in the figure, to ease the visualization) and the dkAEs approximated kernel matrices, relative to test data, are displayed for two different $\lambda$ values. 
Note that, to illustrate the difference to a traditional sAE, one of the two $\lambda$ values is set to zero. 
It can be clearly seen that, for $\lambda=0.1$, both the reconstruction error and kernel matrix closely resemble the original, which agrees with the plots in Figure~\ref{fig:lambda_experiment}.

Inspecting the kernels obtained in Figure~\ref{fig:lambda_experiment_reconstruction}, we compare the distance between the kernel matrices, $\mathbf{C}$ and $\mathbf{P}$, and the ideal kernel matrix, obtained by considering supervised information. We build the ideal kernel matrix $\mathbf{K}_I$, where $K_I(i,j) = 1$ if elements $i$ and $j$ belong to same class, otherwise $K_I(i,j) = 0$. 
Table~\ref{tab:results_ideal} illustrates that the kernel approximation produced by dkAE outperforms a traditional sAE with regards to kernel alignment with the ideal kernel. Additionally, it can be seen that the kernel approximation $\mathbf{C}$ actually is more similar to the ideal kernel than the kernel prior, which we hypothesize is due to the reconstruction objective, which allows the codes to capture additional information (w.r.t. to PCK) about the structure of the input space.

\bgroup
\def\arraystretch{1} 
\setlength\tabcolsep{1em} 
  \begin{table*}[tbp]\scriptsize\centering
  \caption{We compute $L_c$ with respect to the ideal kernel matrix $\mathbf{K}_I$ for our test dataset ($10$ classes) and compare the relative improvement for the three kernels in Figure~\ref{fig:lambda_experiment_reconstruction}. It can be seen that the kernel matrix produced by dkAE ($\mathbf{C}$) is quantitatively comparable to the prior kernel ($\mathbf{P}$) with regards to its distance from the ideal kernel matrix and outperforms the traditional sAE ($\mathbf{K}_{AE}$).}
  \vspace{-0.05cm}
  \begin{tabular}{cccc|c}
    \bottomrule[1.5pt]
    \multirow{2}{*}{\textbf{Kernel}} & \multicolumn{3}{c}{Improvement [\%]~vs.} & \multirow{2}{*}{$L_c(\cdot, \mathbf{K}_I)$} \\
    & $\mathbf{P}$ & $\mathbf{K}_{AE}$ & $\mathbf{C}$ & \\
    \toprule[1.5pt]
    $\mathbf{P}$ & 0 & 12.7  & -0.2 & 1.0132 \\ 
    $\mathbf{K}_{AE}$ & -11.3 & 0 & -11.4 & 1.1417 \\ 
    $\mathbf{C}$ & 0.2 & 12.9 & 0 & 1.0115 \\ 
    \toprule[1.5pt]
  \end{tabular}
  \label{tab:results_ideal}
  \end{table*}
\egroup

\subsection{Approximation of kernel matrix given as prior}
\label{sec:exp2}
In order to quantify the kernel alignment performance, we compare dkAE to the approximation provided by kPCA when varying the number of retained principal components.
For this test, we take the kernel matrix $\mathbf{P}$ of the training set and compute its eigendecomposition. We then select an increasing number of components $m$ (with $m\geq1$ components related to the largest eigenvalues) to project the input data as follows: $\mathbf{Z}_m = \mathbf{E}_m \bm{\Lambda}_{m}^{1/2}, d=2, ..., N$. 
The approximation of the original kernel matrix (prior) is then given by $\mathbf{K}_m = \mathbf{Z}_m \mathbf{Z}_{m}^{T}$. We compute the distance between $\mathbf{K}_m$ and $\mathbf{P}$ following Equation~\ref{eq:normdistance} and compare it to the dissimilarity between $\mathbf{P}$ and $\mathbf{C}$. 
For evaluating the out-of-sample performance, we use the Nystr{\"o}m approximation for kPCA~\cite{scholkopf1998nonlinear} and compare it to the dkAE kernel approximation on the test set.

\begin{figure}[tbp]
  \centering
\includegraphics[width=0.45\textwidth, keepaspectratio, trim={0.7cm 0.1cm 0.7cm 0.1cm},clip]{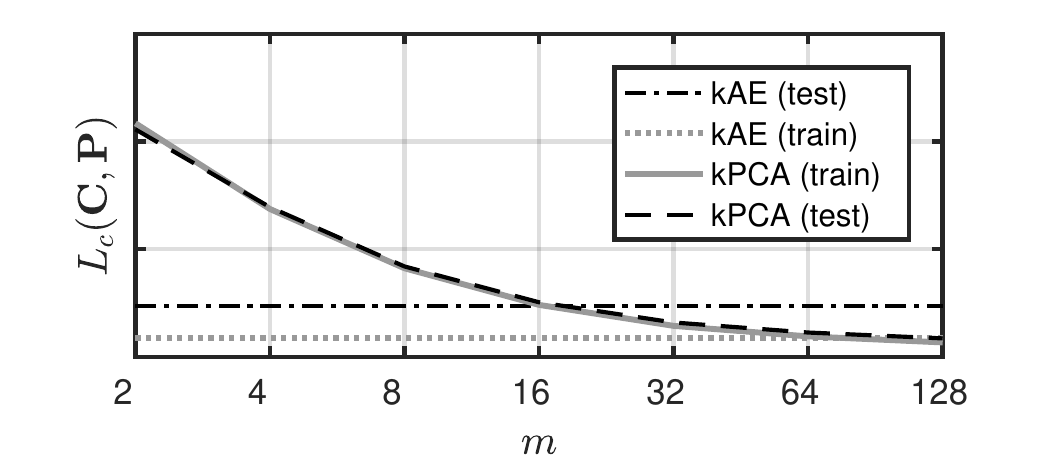}
  \caption{Comparing dkAEs approximation of the kernel matrix to kPCA for an increasing number of components. The plot shows that dkAE reconstruction is more accurate for low number (i.e., $m<16$) of components.}
    \label{fig:kPCA_exp}
\end{figure}

Figure~\ref{fig:kPCA_exp} shows that the approximation obtained by means of dkAEs achieves a more accurate reconstruction then kPCA when using a small number of components, i.e., $m<16$. Note that it is common in spectral methods to chose a number of components equal to the number of classes in the dataset~\cite{ng2001spectral}, in which case, for the $10$ classes in MNIST, dkAE would outperform kPCA. As expected, when the number of selected components increases, the approximation provided by kPCA is better. However, as shown in the previous experiment (Section \ref{sec:exp1b}), this does not mean that the approximation performs better with regards to the ideal kernel. In fact, in that experiment the kernel approximation of dkAE actually performed at least as well as the prior kernel (kPCA with all components taken into account).

\section{Applications of dKAEs in classification, denoising, and visualization of high-dimensional data}
\label{sec:experiments}
In this section, we evaluate the effectiveness of dkAEs learned representations on multiple tasks.
In Section \ref{sec:classificaiton-visualization}, we compare classification performance on different benchmarks and illustrate how dkAEs can be used also for visualization of high-dimensional data. In Section \ref{sec:denoising}, we present an application of our method for image denoising, where we apply PCA in dkAE code space $\mathcal{C}$ to remove noise.

For our classification experiments, apart from MNIST, we consider also the following datasets:
\begin{itemize}
    \item CIFAR-10, which consists of $60000$ $32\times 32$ color images belonging to 10 classes (airplane, automobile, bird, cat, deer, dog, frog, horse, ship, and truck)~\cite{krizhevsky2009learning}. Similar to the MNIST dataset, we consider a subset of $20000$ samples.
    \item Cloud, a dataset containing three multispectral satellite images captured over Spain and France.
    Each pixel in the images is represented by 19 dimensions, where 13 dimensions represent spectral bands
        from the MEdium Resolution Imaging Spectrometer (MERIS) instrument on board the Environmental Satellite (ENVISAT) \cite{rast1999esa},
        while the remaining six dimensions are related to physical features \cite{gomez2007cloud}.
    Each pixel is labeled according to the presence of a cloud in that particular area.
    This is a binary classification task, where the goal is to identify areas in the image which are obscured
        by clouds.
    The dataset is identical to the one previously used in \cite{gomez2012kernel}.
    
    Similar to the MNIST dataset, we consider a subset of $20000$ samples, where the training set consists of pixels sampled from
        one image, the validation set is sampled from a different image and the test set is sampled from the remaining image.
    \item Reuters, which consists of $800000$ news stories that have been manually categorized into a category tree~\cite{lewis2004rcv1}. Similar to~\cite{xie2015unsupervised} we choose the four root categories as labels and remove stories that are labeled with multiple root categories.
    To represent each news story we compute feature vectors consisting of the Term Frequency-Inverse Document Frequency (TF-IDF) of the 2000 most frequently occurring word stems and then use a Singular Value Decomposition (SVD) to produce $20$ dimensional vectors prior to training.
    The SVD is performed on the training set, with out-of-sample transformations for validation and test sets.
\end{itemize}

\subsection{Visualization and classification in code space}
\label{sec:classificaiton-visualization}

In order to evaluate the learned representation and illustrate the use of our method on an independent test set, we evaluate the classification performance of the learned representation. Here we make use of a linear support vector machine (SVM) operating in the code space and compare it to a linear and a non-linear kernel SVM (kSVM) operating directly in input space. 
The dKAE is trained on the training dataset, the SVMs model parameters are optimized on the validation set and the final accuracy is shown on the test dataset. Table~\ref{tab:results_ideal} shows that linear SVM trained in the code space (cSVM) outperforms the SVM models operating in input space on all datasets.

As a consequence of the fact that our code representation is controlled by an arbitrary kernel matrix, we can also extend our work to learn representations in a supervised manner by aligning the code matrix $\mathbf C$ with the ideal kernel matrix. Similar to the experiments for the unsupervised representation, we train a linear SVM in the code space representation that has been learned (by exploiting supervised information) and provide the achieved accuracy (scSVM) in Table~\ref{tab:results_class}. As expected, when exploiting supervised information to learn representations, improvements are observed for all datasets.
To illustrate the robustness of our approach with respect to architectural choices, we make use of the same architecture for all datasets, namely the one described in Section~\ref{sec:analysis}. Note, however, to avoid overfitting to the training data when training the supervised representation for the CLOUD dataset, the architecture for this particular dataset was reduced to $d-50-50-200-200$ for all experiments.

\bgroup
\def\arraystretch{1} 
\setlength\tabcolsep{1em} 
  \begin{table*}[tbp!]\scriptsize\centering
  \caption{Quantitative analysis of the learned feature representation of dkAE for classification tasks. A linear SVM operating in code space (cSVM) is compared with a linear SVM and a kernel SVM (kSVM) operating directly in input space. We also considered a linear SVM operating in code space where the prior $\mathbf{P}$ for the alignment is given by the outer product of class labels (scSVM).}
  \vspace{-0.05cm}
  \begin{tabular}{c|cccc}
    \bottomrule[1.5pt]
    Method & MNIST & CLOUD & CIFAR-10 & REUTERS\\
    \toprule[1.5pt]
    SVM & 90.60 & 99.50 & 36.60 & 91.40\\
    kSVM & 93.80 & 99.60 & 36.93 & 93.23 \\ 
    cSVM & 94.80 & 99.63 & 38.17 & 93.77 \\
    scSVM & 96.23 & 99.70 & 42.73 & 94.17 \\
    \toprule[1.5pt]
  \end{tabular}
  \label{tab:results_class}
  \end{table*}
\egroup

\begin{figure}[tbp!]
\centering
\minipage[t]{0.4\textwidth}%
  \fbox{\includegraphics[height=0.6\linewidth,width=0.8\linewidth]{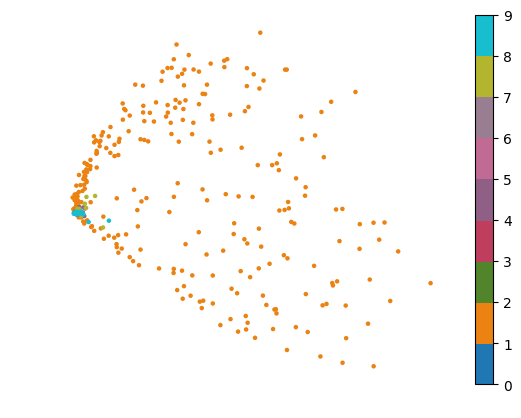}}
  \centering{KECA}
\endminipage\hspace{0.2cm}
\minipage[t]{0.4\textwidth}%
  \fbox{\includegraphics[height=0.6\linewidth,width=0.8\linewidth]{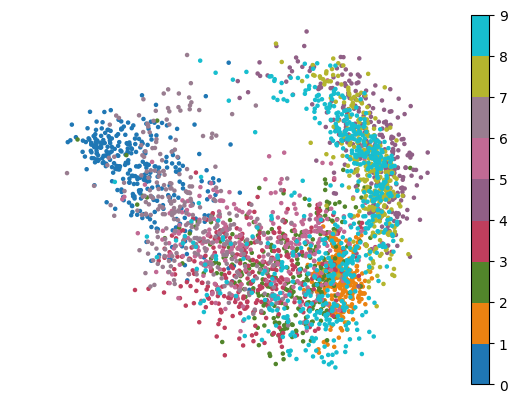}}
  \centering{ISOMAP}
\endminipage\vspace{0.2cm}
\minipage[t]{0.4\textwidth}
  \fbox{\includegraphics[height=0.6\linewidth,width=0.8\linewidth]{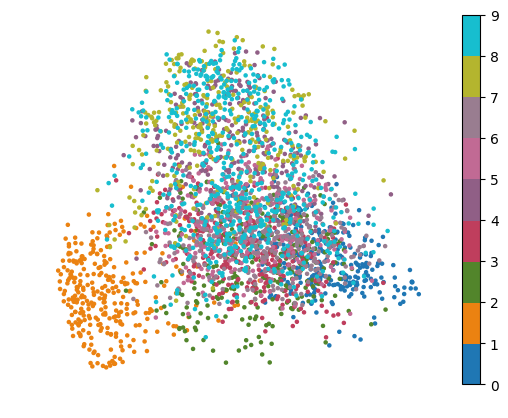}}
  \centering{AE+PCA}
\endminipage\hspace{0.2cm}
\minipage[t]{0.4\textwidth}
  \fbox{\includegraphics[height=0.6\linewidth,width=0.8\linewidth]{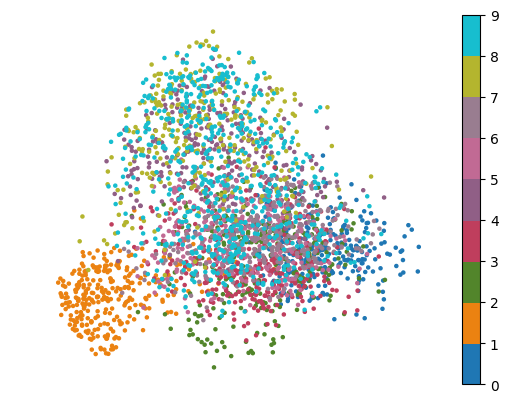}}
  \centering{DAE+PCA}
\endminipage\vspace{0.2cm}
\minipage[t]{0.4\textwidth}
  \fbox{\includegraphics[height=0.6\linewidth,width=0.8\linewidth]{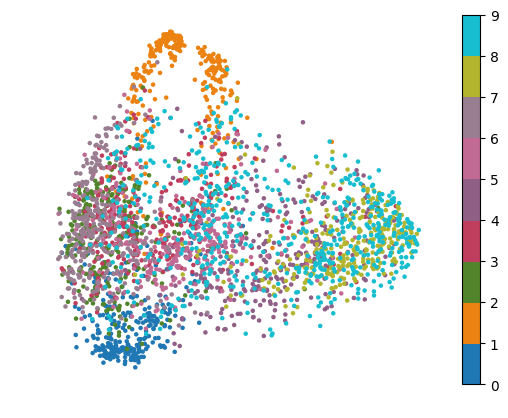}}
  \centering{dKAE+PCA}
\endminipage
\caption{MNIST data. Two dimensional embedding of the code space obtained using standard AEs, DAEs and our dKAE. The codes are projected to two dimensions using PCA. We compare their preformance to non-linear dimensionality reduction techniques KECA and ISOMAP.}
\label{fig:visualization}
\end{figure}

Now we assess the capability to visualize high-dimensional data.
Figure~\ref{fig:visualization} shows the visualization of a low-dimensional representation learned by dkAE for the MNIST dataset; here, we consider 2000-dimensional codes. We utilize PCA to map the learned codes to two-dimensional vectors.
We take into account also the low-dimensional representation learned by four alternative methods, namely an autoencoder without the use of kernel alignment, a denoising autoencoder (DAE)~\cite{vincent2010stacked} with 20\% masking noise, and kernel entropy component analysis (KECA)~\cite{jenssen2010kernel} as well as ISOMAP \cite{tenenbaum2000global}, two popular non-linear dimensionality reduction methods. Note, that the visualization here is presented for the test set and not the training data. For KECA we utilize an RBF kernel with $\sigma$ being set to 15 percent of the median pairwise euclidean distances between datapoints, following a rule of thumb from~\cite{jenssen2010kernel}. We use KECA to reduce the dimensionality to $10$ dimensions, the number classes in the dataset, before using PCA to reduce it further down to $2$.
In order to provide a quantitative evaluation of the visualizations, we consider the generalization error on a 1-Nearest Neighbor classification task following the example of~\cite{sanguinetti2008dimensionality}. Results are shown in Table \ref{tab:visualization}, which demonstrate the superior performance obtained by means of dkAE.


\bgroup
\def\arraystretch{1.4} 
\setlength\tabcolsep{0.2em} 
  \begin{table}[tbp!]\scriptsize\centering
  \caption{1-nearest neighbor classification accuracy on representations shown in Figure \ref{fig:visualization}. The overall best result is highlighted in bold.}
  \vspace{-0.05cm}
  \begin{tabular}{ccccc}
    \bottomrule[1.5pt]
    \textbf{KECA} & \textbf{ISOMAP} & \textbf{AE+PCA} & \textbf{DAE+PCA} & \textbf{dkAE+PCA}\\
    \toprule[1.5pt]
    29.5 & 36.8 & 30.5 & 31.2 & \bf{39.6}\\
    \toprule[1.5pt]
  \end{tabular}
  \label{tab:visualization}
  \end{table}
\egroup

\subsection{Denoising and visualizing code space traversal in input space}
\label{sec:denoising}
Here, we highlight the potential of performing explicit operations in code space as initially described in Section~\ref{sec:lin_op}.
We try to emulate kPCA by performing PCA in our learned code space and evaluate the performance on a denoising task.
Denoising is a task that requires both a mapping to the kernel space, as well as a back-projection to the input space.
Traditional kernel methods cannot perform back-projection explicitly; approximate solutions have been proposed in the literature~\cite{bakir2004learning,honeine2011closed}. We choose the method proposed by Bakir et al.~\cite{bakir2004learning}, where they use kernel ridge regression, such that a different kernel (in our case an RBF) can be used for back-mapping.
Due to the challenge of finding a good $\sigma$ for the RBF kernel that works on all MNIST numbers, we performed this test on the $5$ and $6$ class only. The regularization parameter and the $\sigma$ required for the back-projection where found via grid search, where the best regularization parameter according to mean squared error (MSE) reconstruction was found to be $0.5$ and $\sigma$ as the median of the Euclidean distances between the projected feature vectors.

\begin{figure}[tbp!]
\centering
\minipage[t]{0.23\textwidth}
  \includegraphics[width=\linewidth]{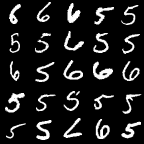}
  \centering{Original}
\endminipage\hspace{0.2cm}
\minipage[t]{0.23\textwidth}%
  \includegraphics[width=\linewidth]{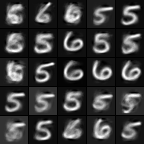}
  \centering{kPCA}
\endminipage\hspace{0.2cm}
\minipage[t]{0.23\textwidth}
  \includegraphics[width=\linewidth]{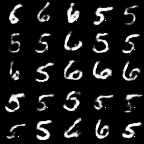}
  \centering{DAE+PCA}
\endminipage\hspace{0.2cm}
\minipage[t]{0.23\textwidth}
  \includegraphics[width=\linewidth]{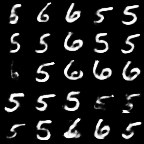}
  \centering{dKAE+PCA}
\endminipage
\caption{Denoising with kPCA in input space and PCA in code space.}
\label{fig:denoising_res}
\end{figure}

Both models are fitted on the training set and additive Gaussian noise is added to the test set. For both methods, $32$ principal components are used. Table~\ref{tab:denoising_res} shows that dkAE+PCA outperforms kPCAs reconstruction in terms of MSE. However, as MSE is not necessarily a good measure for denoising~\cite{bakir2004learning}, we also visualize the results in Figure~\ref{fig:denoising_res}. It can be seen that dkAE yields sharper images in the denoising task. We further compare the results to a denoising autoencoder (DAE+PCA). We observe that the denoising autoencoder is able to outperform the dkAE with regards to the MSE measure as it is explicitly trained for the denoising task. Qualitatively, however, we observe in Figure~\ref{fig:denoising_res} that the qualitative difference between these two is small, with DAE outperforming the dkAE on some images while producing more washed out images on others. For example, the reconstruction of the first image in the first row is better reconstructed using the DAE, while the second and fifth image in the first row are better reconstructed by the dKAE.

\begin{figure}[tbp!]
\centering
\minipage[t]{0.3\textwidth}
  \includegraphics[width=\linewidth,trim={0cm 1cm 0 0},clip]{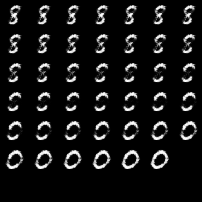}
\endminipage\hspace{0.2cm}
\minipage[t]{0.3\textwidth}%
  \includegraphics[width=\linewidth,trim={0cm 1cm 0 0},clip]{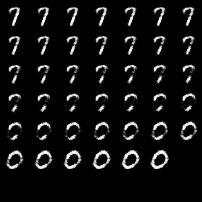}
\endminipage\hspace{0.2cm}
\minipage[t]{0.3\textwidth}
  \includegraphics[width=\linewidth,trim={0cm 1cm 0 0},clip]{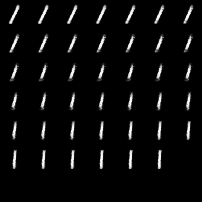}
\endminipage
\caption{First and the last image of each panel show two $k$-means centroids in code space obtained on the MNIST dataset. Additional numbers are generated by ``walking'' on interpolated points between the two centroids.}\label{fig:walk}
\end{figure}

\bgroup
\scriptsize\centering
\def\arraystretch{1.4} 
\setlength\tabcolsep{1em} 
\begin{table}
\caption{MSE of reconstruction.}%
  \begin{tabular}{cccc}
    \bottomrule[1.5pt]
    \textbf{Noise std.} & \textbf{kPCA} & \textbf{DAE+PCA} & \textbf{dkAE+PCA}\\
    \toprule[1.5pt]
     0.25 & 0.0427 & 0.0173 & 0.0358  \\ 
    \toprule[1.5pt]
  \end{tabular}
  \label{tab:denoising_res}
  \end{table}
\egroup

dkAE allows to explicitly explore the code space beyond the image of the dataset at hand and accordingly generate new instances with related representations in input space.
To this end, we visualize the effect of movements in code space, illustrated in Figure~\ref{fig:walk}. For this experiment, we perform $k$-means clustering in code space and chose the number of clusters to be equal to the number of classes in MNIST. We select pairs of cluster centroids at random and interpolate between two centroids following a straight path in code space; in future works, we will consider also non-linear methods to obtain a smoother interpolation between the centroids~\cite{shao2017riemannian}.
The first and last image in Figure~\ref{fig:walk} correspond to the cluster centers. The intermediate images are generated by mapping points along the aforementioned path in code space back to the input space by means of the trained decoder. In the first two panels, we observe a smooth transition of an $8$ and a $7$ to a $0$. The third panel, instead, illustrates that $k$-means found two clusters in the $1$s class, one for the far leaning ones and one for the straight ones. Interpolating between these two allows us to generate numbers with a varying degree of leaning to the right.

\section{Conclusions}
\label{sec:conclusions}

We proposed a novel model for autoencoders, dubbed deep kernelized autoencoders, that exploits information provided by a user-defined kernel matrix to learn similarity-preserving data representations. The proposed model is trained end-to-end in an unsupervised way.
By means of a parameter-free kernel alignment procedure based on inner products between codes, we are able to approximate arbitrary kernel functions defined in input space. This allows us to learn an explicit mapping from the input space to the code space, as well as the backward mapping.
We evaluated the learned data representations on classification tasks and illustrated how the learned backmapping can be used to visualize operations performed directly in code space. In addition, the proposed autoencoder enables us to emulate well-known kernel methods for unsupervised learning, such as kernel PCA; however, our approach scales well with the number of data points as it is not based on eigendecomposition procedures.

In future work, we will continue to investigate this line of research by exploring alternative loss functions for kernel alignment, beyond those based on Frobenius norm. In particular, we will investigate the use of information-theoretic divergence measures and formulations based on mutual information between positive semi-definite matrices.

\section*{Acknowledgment}
We gratefully acknowledge the support of NVIDIA Corporation with the donation of the GPU used for this research. This work was partially funded by the Norwegian Research Council FRIPRO grant no.\ 239844 on developing the \emph{Next Generation Learning Machines}.

\section*{References}
\bibliographystyle{abbrvnat}
\bibliography{bibliography.bib}

\end{document}